\pdfoutput=1

\documentclass[11pt]{article}

\usepackage[final]{acl}

\usepackage{times}
\usepackage{latexsym}

\usepackage[T1]{fontenc}

\usepackage[utf8]{inputenc}

\usepackage{microtype}
\usepackage{lipsum} 
\usepackage{booktabs}
\usepackage{graphicx}
\usepackage{amsmath}
\usepackage[framemethod=tikz]{mdframed}
\usepackage{comment}
\usepackage{multirow}
\usepackage{hyperref}
\usepackage{adjustbox}
\usepackage{footmisc}
\usepackage[misc]{ifsym}
\usepackage{subcaption}
\usepackage{extarrows}
\usepackage{amsmath}
\usepackage{mathrsfs}
\usepackage{microtype}
\usepackage{color}
\usepackage{colortbl} 
\usepackage{mathrsfs}
\usepackage{float}

\usepackage{array}
\usepackage{nicematrix}
\usepackage{longtable}
\usepackage{rotating}
\usepackage{pdflscape}

\title{
An Empirical Study of Multitask Learning to Improve Open Domain Dialogue Systems
}

\author{Mehrdad Farahani\textsuperscript{\normalfont1}\hspace{5mm}Richard Johansson\textsuperscript{\normalfont1,2} \\
  \textsuperscript{1}Chalmers University of Technology, 
  \textsuperscript{2}University of Gothenburg\\
  \texttt{\{mehrdad.farahani, richajo\}@chalmers.se}}

\begin{document}
\maketitle
\begin{abstract}
Autoregressive models used to generate responses in open-domain dialogue systems often struggle to take long-term context into account and to maintain consistency over a dialogue.
Previous research in open-domain dialogue generation has shown that the use of \emph{auxiliary tasks} can introduce inductive biases that encourage the model to improve these qualities. However, most previous research has focused on encoder-only or encoder/decoder models, while the use of auxiliary tasks in \emph{decoder-only} autoregressive models is under-explored.
This paper describes an investigation where four different auxiliary tasks are added to small and medium-sized GPT-2 models fine-tuned on the PersonaChat and DailyDialog datasets.
The results show that the introduction of the new auxiliary tasks leads to small but consistent improvement in evaluations of the investigated models.
\end{abstract}

\section{Introduction}

In recent years, open-domain dialogue systems have experienced increased research due to the availability of large corpora of dialogue and advances in deep learning techniques \cite{gao_etal_2018}. Unlike task-oriented dialogue systems designed for specific domains or tasks, such as flight booking, hotel reservation, customer service, and technical support \cite{budzianowski_&_vulic_2019,budzianowski_etal_2018,chao_&_lane_2019}, open-domain dialogue systems aim to have long-term connections with users by satisfying their emotional, social, and communication needs. Therefore, such a system must comprehend the dialogue context and user demands in order to select the appropriate skill at the appropriate time and generate consistent \cite{li_etal_2016b,luan_etal_2017} and grounded \cite{ghaznavinejad_etal_2018,moon_etal_2019} interpersonal responses. Open-domain dialogue systems can include single-turn or multi-turn dialogues, where the context and topic of the conversation may change throughout the interaction \cite{dinan_etal_2019}.

Multi-turn open-domain dialogue systems need to maintain the context of the conversation, generate appropriate responses concerning the context and predefined characteristics (persona), and handle various forms of input. A persona is a set of characteristics or attributes that describe a virtual agent's background, personality, and behavior. These attributes include the agent's name, age, gender, profession, interests, and other aspects \cite{zhang_etal_2018}, while a conversation context refers to the background information or previous interactions relevant to the current conversation, with word-level and utterance-level dependencies \cite{zhao_etal_2020}.

Recent developments in transformer-based architectures \cite{vaswani_etal_2017,raffel_etal_2019,lewis_etal_2020} for large-scale pre-training, such as OpenAI's GPT-2 \cite{radford_etal_2019}, have shown exceptional results.
Later, the models are fine-tuned via more technical steps and at different scales on a large-scale dialogue corpus \cite{zhang_etal_2019,freitas_etal_2020,thoppilan_etal_2022,shuster_etal_2022}.

Pre-trained models on conversational datasets typically process dialogue context (a list of utterances) as a sequence of tokens per utterance to generate responses. Although these approaches show effective results compared to previous approaches, they still need to catch the latent information in more complex structures rather than just tokens \cite{gu_etal_2020,zhao_etal_2020}. A conversational domain is distinguished by the presence of another component called utterances\footnote{An utterance is a spoken or written sentence or phrase that is used to convey meaning or participate in a dialogue.} that plays an imperative role in conveying higher-level information in addition to tokens and their local relationships. Recent research has put forth the use of \emph{auxiliary tasks} as a form of regularization during the fine-tuning of models as a means to address the aforementioned issue. A majority of these additional training objectives are implemented solely on the encoder-only and encoder-decoder architectures.

However, while the use of auxiliary tasks has led to improvement in encoder-only and encoder/decoder models, 
recent work has not explored the application of auxiliary tasks in \emph{decoder-only} models. In this research, we propose incorporating auxiliary tasks on top of an autoregressive decoder-only model to examine and enhance the quality of generated responses concerning the latent information present within utterances. Additionally, we demonstrated the impact of various auxiliary tasks on distinct elements of dialogue across two benchmark datasets. By examining the effect of different auxiliary tasks on various components of dialogue, we aimed to provide a deeper understanding of how these tasks can influence the performance and outcomes of conversational systems. Additionally associated code to this research can be found in our GitHub repository.\footnote{\href{https://github.com/m3hrdadfi/MoGPT}{https://github.com/m3hrdadfi/MoGPT}}

\section{Related Works}

The motivation for this research is drawn from recent investigations into the utilization of auxiliary tasks to enhance the generated responses in open-domain dialogue systems by considering the context. To this end, we present and analyze these recent studies in this section. Previous studies in this field can be broadly classified into three general categories. The first category pertains to the widespread use of encoder-decoder models in dialogue response generation, which have been observed to produce generic and uninteresting responses (e.g., "I'm good", "I don't know"). \newcite{zhao_etal_2020} proposed an encoder-decoder architecture with two auxiliary tasks at token and utterance levels that can effectively exploit conversation context to generate responses, including order recovery and masked context recovery. Analogously, \newcite{mehri_etal_2019} examined a range of unsupervised pre-training objectives for acquiring dialogue context representations via encoder-decoder models by incorporating four auxiliary tasks, including next-utterance retrieval, next-utterance generation, masked-utterance retrieval, and inconsistency identification.

DialogBERT is a unique design that employs a hierarchical Transformer architecture to comprehensively capture the context of dialogue \cite{gu_etal_2020}. Using two training objectives, similar to BERT \cite{devlin_etal_2019}, allows the model to understand a conversation's nuances effectively. In the first objective, masked context regression, the model is trained to predict the missing context from a dialogue, and in the second objective, distributed utterance order prediction, the model is trained to predict the order of spoken utterances in a conversation so that it understands the flow and logic.

Lastly, decoder-only models, like DialoGPT \cite{zhang_etal_2019}, make use of only the final component of the encoder-decoder structure. DialoGPT in particular, extends the GPT-2 \cite{radford_etal_2019} architecture by being specifically developed and trained on a large corpus of dialogue data to generate responses in a conversational context. However, despite its ability to perform well in single-turn conversation, its lack of capability to capture latent information behind utterances in a multi-turn conversation, results in an inadequate understanding of the context.  The utilization of auxiliary tasks in decoder-only models is a well-established practice. For instance, the GPT-2 based model TransferTransfo \cite{wolf_etal_219}, which adopts a multi-task objective, showed improvement over the basic GPT-2. These auxiliary tasks primarily take the form of sequence classification tasks.

\section{Method}

\subsection{A Problem Definition}
\begin{figure*}[!ht]
    \centering
    \includegraphics[height=4in]{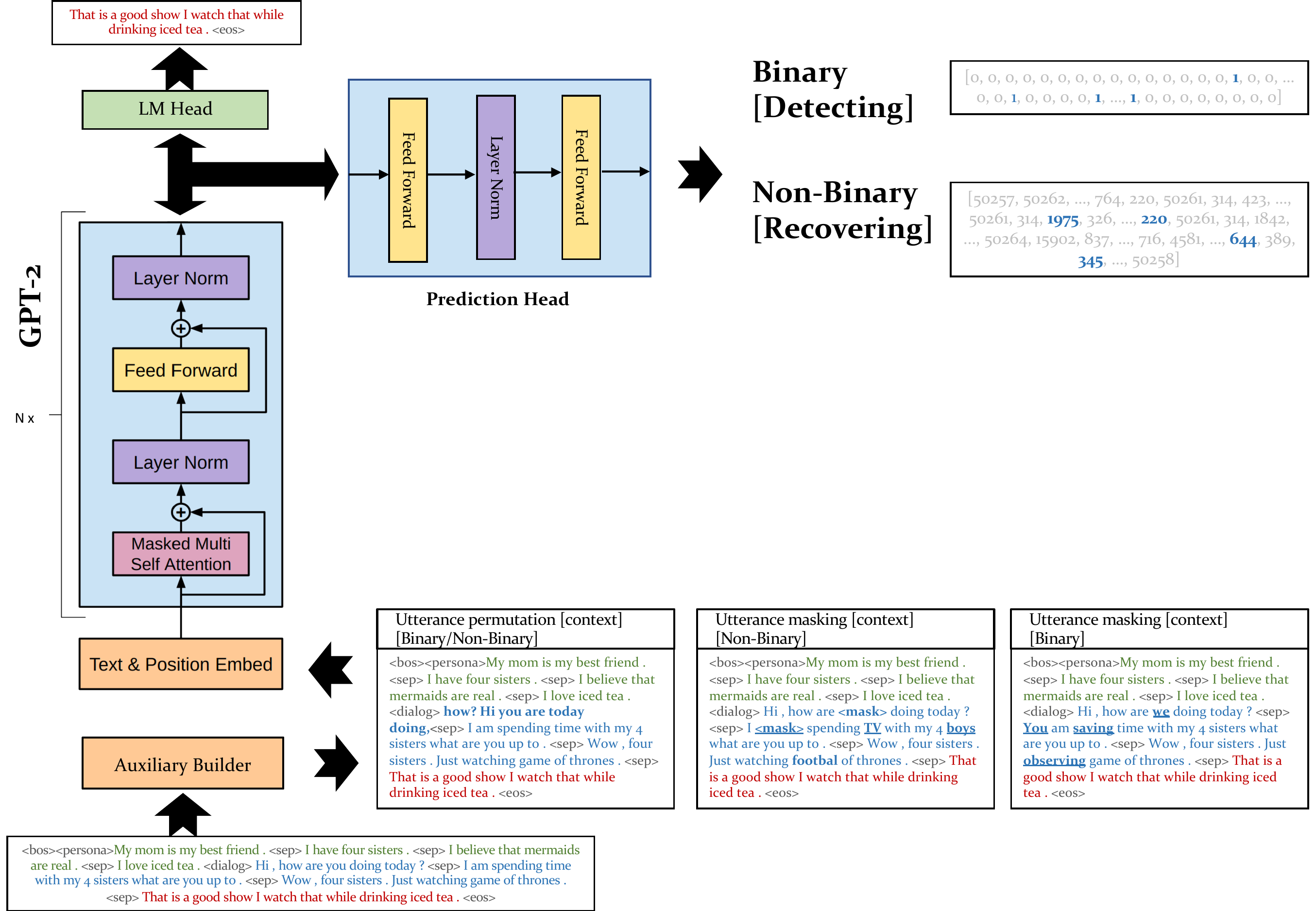}
    \caption{
        \label{figure:model_struc}
        This figure illustrates the auxiliary tasks and the proposed model. The input to the model (our prompt) includes a combination of \textbf{persona}, \textbf{context}, and the \textbf{last conversation}. Each component is separated into special tokens and preceded by a unique token that signifies its component. The model's objective (LM objective) is to generate the final conversational component of the agent's response while disregarding any prior parts.
    }
\end{figure*}

In this section, the necessary notations utilized are presented, and the learned tasks are briefly outlined. 
Let $d^{(i)} = \left(p_1, p_2, \ldots, p_N, u_1, u_2, \ldots, u_T\right)$ denote the $i$-th dialogue session in the dataset $\mathcal{D}$, where $\mathcal{C}=$ $\left(u_1, u_2, \ldots, u_{T-1}\right)$ is the dialogue context (history), $\mathcal{P}=$ $\left(p_1, p_2, \ldots, p_{N}\right)$ is the dialogue persona (personality of the system) and $u_T$ is the response regarding to the persona and the context. Each $u_i=\left(w_1^i, w_2^i, \ldots, w_{\left|u_i\right|}^i\right)$ in $\mathcal{C}$ is an utterance and $w_j^i$ is the $j$-th word in $u_i$. Then, we aim to generate contextually relevant responses for multi-turn conversations using self-supervised auxiliary tasks. Our approach involves two major components, a language model trained based on the GPT-2 and a classification model on top of the GPT-2 used for auxiliary parts. This simple structure has been found to be effective in producing consistent responses. As such, two auxiliary tasks have been designed over language modeling (LM) to improve the system's performance further. Order and masked recovery tasks are designed to enhance the self-attention module's capacity to capture linguistic affinities. The utterance permutation task enhances the self-attention module's ability to grasp word and utterance sequences, while the masking task seeks to reinforce semantic connections between words and utterances by optimizing the self-attention mechanism. These auxiliary tasks are critical in providing additional supervision signals to the model, leading to improved language modeling performance. Figure \ref{figure:model_struc} illustrates the model. Lastly, a total loss function is defined to incorporate these auxiliary tasks and the primary objective of language modeling. It serves as the optimization target during training and guides the model toward producing accurate and consistent responses.

\begin{equation}
\mathcal{L}=\mathcal{L}_{\text{LM}}+\alpha \mathcal{L}_{\text{aux}}
\end{equation}
Here, $\alpha$ is a hyper-parameter that controls the trade-off between LM and the objectives of the auxiliary tasks. \\

\subsection{Auxiliary Tasks}

Recent research \cite{sankar_etal_2019} has shown that Transformer-based autoregressive models are robust to unrealistic perturbations at both the utterance and word levels. However, despite this robustness, the study suggests that these models have learned a bag-of-words-like representation rather than genuinely understanding language structure and meaning. On the other hand, understanding context is crucial to producing coherent and consistent responses in open-domain dialogue systems. While the connections between words within an individual utterance are essential for determining the meaning, it is also necessary to consider the relationships between utterances to fully understand the context of the conversation. To enhance the language model's comprehension and its ability to generate accurate and consistent outputs, it was deemed necessary to provide additional means to understand the relationships between the order of utterances and their meaning and to capture the sequential structure of language, as well as to comprehend the relationships between individual words in an utterance to grasp the semantic structure of language. We propose two auxiliary tasks for this purpose in this paper.

\subsubsection{Utterance Permutation (UP)}

In order to retain the sequential structure of language, re-ordering utterances is defined as an auxiliary generator in two ways: detection or recovering methods by rearranging 10\% of utterances in a dialog chosen by 15\% of all dialogues in the collection. Depending on the dataset, this task can be implemented based on a persona, context (history), or both. 


In this work, we considered two approaches to implementing UP as auxiliary tasks:
\begin{itemize}
\item \emph{detecting} (UPD), 
implemented as a binary token classification task.
\item \emph{recovering} (UPR), implemented as a non-binary token classification task.
\end{itemize}
In UPR, we attempt to predict the correct tokens regarding the re-ordered tokens; in UPD, we only determine whether or not the tokens are in the right place.

\subsubsection{Utterance Masking (UM)}
In our effort to comprehend the semantic structure underlying utterances, we devised the utterance masking task. This task is executed using two distinct approaches, analogously to the two methods described above:
\begin{itemize}
\item \emph{detecting} the tokens in the masked utterances (UMD), implemented as a binary classification task.
\item \emph{recovering} the tokens in the masked utterance (UMR).
\end{itemize}
In both methods, 15\% of the tokens within each dialogue are selected, with 80\% of these tokens being replaced in the non-binary approach by the <mask> token and by synonyms in the binary approach. In the non-binary method, 10\% of the tokens were randomly substituted from the dictionary, while in the binary approach, they are replaced with antonyms. The final 10\% of tokens were preserved in their original form.

\section{Dataset and Experiments}
The following section provides detail on the dataset and experimental settings used in our experiments.

\subsection{Dataset}
The experiments in this paper are conducted using two benchmark datasets for open-domain dialogue generation, PersonaChat \cite{zhang_etal_2018} and DailyDialog \cite{li_etal_2017}. PersonaChat is a large-scale dataset collected by encouraging two individuals to engage in open-domain conversations while exchanging personal information to create personas. The dataset contains over 163,064 utterances (11,907 dialogues) for training and 15,0264 utterances (968 dialogues) for testing. The conversations are naturally diverse, covering various topics and perspectives. In addition, the personal information provided allows the model to generate more informed and coherent responses due to predefined personalities. A true-cased version of PersonaChat\footnote{\href{https://huggingface.co/datasets/bavard/personachat_truecased}{bavard\slash personachat\_truecased}} is used in the experiments to maintain consistency with the other datasets.
On the other hand, the DailyDialog is a small dataset consisting of 13,118 multi-turn dialogues collected from various daily situations. Both datasets are pre-processed to ensure that all conversations are well-formed and coherent and that the data is in a suitable format for training and the auxiliary generator. The datasets are split into training, validation, and test sets for experimentation.

\subsection{Baselines}
We compared our approach to DialoGPT \cite{zhang_etal_2019}, a Transformer-based response generation model. We design a new prompt that is suitable for our auxiliary tasks. In order to ensure a fair comparison, we fine-tune the GPT-2 model on both of the two multi-turn datasets with the new prompt and using the same configurations introduced by DialoGPT (also known as VanillaGPT-2). This allows us to compare our approach to DialoGPT under the same conditions.

\subsection{Implementation Details}
Our implementation of both approaches is carried out using PyTorch Lightning\footnote{\href{https://www.pytorchlightning.ai/}{https://www.pytorchlightning.ai/}} and Huggingface Transformers.\footnote{\href{https://huggingface.co/}{https://huggingface.co/}} We train the baseline and our approach on two GPT-2 scales (small 124M and medium 354M parameters). Our approach depends on the dataset we implement the auxiliary tasks on different components persona, context, persona, and context, and by random. All the models are optimized with the AdamW optimizer \cite{loshchilov_etal_2017} using an initial learning rate of 5e-5 and 3e-5, respectively, for the DailyDialog and PersonaChat datasets, and by using the adaptive learning rate scheduler with 5,000 warm-up steps and weight decay of 0.001. Experiments are performed on NVIDIA A100 for five epochs and a different range of hyperparameters regarding the auxiliary tasks, as seen in Table \ref{table:hyper_params}.

\begin{table}[!htbp]
    \centering
      \scriptsize
      \scalebox{0.9}{
        \begin{tabular}{clcccc}
            \toprule
            Auxiliary Task  & $\alpha$ & $P_{do}$ & $P_{reordered}$ & $P_{masked}$ & $P_{changed}$ \\
            \midrule
            
            \multirow{1}{*}{UPD} & 3.0 & 0.15 & 0.1 & - & - \\
            \multirow{1}{*}{UPR} & 1.0 & 0.15 & 0.1 & - & - \\
            \multirow{1}{*}{UMD} & 3.0 & 0.15 & - & 0.8 & 0.5 \\
            \multirow{1}{*}{UMR} & 1.0 & 0.15 & - & 0.8 & 0.5 \\

            \bottomrule
        \end{tabular}
        }
        \caption{
            \label{table:hyper_params}
            Hyperparameters used in the experiments.
        }
\end{table}

\begin{table*}[!ht]
\centering
\scriptsize
\scalebox{0.75}{
\large
\begin{tabular}{llcccccccc}
\multicolumn{9}{c}{DailyDialog}\\
\toprule
Scale & Model & PPL & BLEU & ROUGE-L & BERTScore & MoverScore & Average & Greedy & Extrema \\ \hline

\multirow{6}{*}{\rotatebox[origin=c]{90}{SMALL}} & &  &  &  &  &  & &  &  \\
& VanillaGPT-2 & 11.463 & 1.188 & 0.187 & 0.885 & 0.045 & \cellcolor{blue!20}0.875 & 0.737 & 0.881 \\
& UPD [context] & \cellcolor{blue!20}11.445 & 1.108 & 0.186 & 0.884 & 0.042 & 0.873 & 0.735 & 0.877 \\
& UPR [context] & 11.607 & 0.984 & 0.179 & 0.883 & 0.038 & 0.870 & 0.734 & 0.880 \\
& UMD [context] & 11.484 & \cellcolor{blue!20}1.365 & \cellcolor{blue!20}0.188 & 0.885 & \cellcolor{blue!20}0.047 & 0.870 & \cellcolor{blue!20}0.736 & 0.881 \\
& UMR [context] & 11.859 & 0.999 & 0.184 & 0.884 & 0.040 & 0.871 & 0.735 & 0.879 \\\hline
\multirow{6}{*}{\rotatebox[origin=c]{90}{MEDIUM}} & &  &  &  &  &  & &  &  \\
& VanillaGPT-2 & 10.344 & \cellcolor{blue!20}2.603 & 0.208 & 0.889 & 0.072 & 0.880 & 0.743 & 0.881\\ 
& UPD [context] & \cellcolor{blue!20}9.958 & 2.393 & 0.203 & 0.888 & 0.064 & \cellcolor{blue!20}0.881 & 0.745 & \cellcolor{blue!20}0.883 \\
& UPR [context] & 10.192 & 2.068 & 0.199 & 0.887 & 0.060 & 0.877 & 0.739 & 0.882 \\
& UMD [context] & 10.659 & 2.458 & 0.208 & 0.889 & \cellcolor{blue!20}0.075 & 0.879 & 0.745 & 0.882 \\
& UMR [context] & 10.554 & 1.886 & 0.195 & 0.886 & 0.055 & 0.874 & 0.740 & 0.880 \\

\bottomrule\\
\multicolumn{9}{c}{PERSONA-CHAT}\\
\toprule
Scale & Model & PPL & BLEU & ROUGE-L & BERTScore & MoverScore & Average & Greedy & Extrema \\ \hline

\multirow{20}{*}{\rotatebox[origin=c]{90}{SMALL}} & &  &  &  &  &  & &  &  \\
& VanillaGPT-2 & 13.149 & 1.489 & 0.099 & 0.879 & 0.056 & 0.878 & 0.694 & 0.872 \\
& UPD [persona] & 13.100 & 1.545 & 0.098 & 0.878 & 0.055 & 0.878 & 0.694 & 0.872 \\
& UPD [context] & 13.101 & 1.537 & 0.098 & 0.879 & 0.056 & 0.878 & 0.693 & 0.872 \\
& UPD [persona+context] & 13.089 & 1.426 & 0.097 & 0.878 & 0.054 & 0.877 & 0.693 & 0.872 \\
& UPD [random] & 13.108 & 1.552 & 0.097 & 0.878 & 0.055 & 0.877 & 0.693 & 0.872 \\
& UPR [persona] & 13.111 & 1.489 & 0.097 & 0.879 & 0.055 & 0.878 & 0.693 & 0.872 \\
& UPR [context] & 13.132 & 1.431 & 0.096 & 0.878 & 0.055 & 0.878 & 0.693 & 0.872 \\
& UPR [persona+context] & 13.125 & \cellcolor{blue!20}1.586 & 0.097 & 0.879 & 0.055 & 0.878 & 0.694 & 0.872\\
& UPR [random] & 13.128 & 1.427 & 0.097 & 0.878 & 0.054 & 0.877 & 0.694 & 0.872\\
& UMD [persona] & 13.073 & 1.393 & 0.098 & 0.878 & 0.055 & 0.878 & 0.693 & 0.873 \\
& UMD [context] & 13.126 & 1.538 & 0.099 & 0.879 & 0.056 & 0.878 & 0.694 & 0.873 \\
& UMD [persona+context] & 13.079 & 1.504 & 0.097 & 0.878 & 0.054 & 0.878 & 0.692 & 0.872 \\
& UMD [random] & \cellcolor{blue!20}13.055 & 1.423 & 0.096 & 0.878 & 0.055 & 0.877 & 0.693 & 0.872 \\
& UMR [persona] & 13.309 & 1.488 & 0.097 & 0.878 & 0.055 & 0.878 & 0.693 & 0.872 \\
& UMR [context] & 13.265 & 1.459 & 0.098 & 0.879 & 0.055 & 0.878 & 0.694 & 0.872 \\
& UMR [persona+context] & 13.362 & 1.371 & 0.096 & 0.878 & 0.053 & 0.878 & 0.693 & 0.872 \\
& UMR [random] & 13.263 & 1.454 & 0.098 & 0.878 & 0.055 & 0.878 & 0.694 & 0.872\\\\

\multirow{20}{*}{\rotatebox[origin=c]{90}{MEDIUM}} & &  &  &  &  &  & &  &  \\
& VanillaGPT-2 & 10.975 & 1.657 & 0.100 & 0.879 & 0.060 & 0.878 & 0.695 & 0.873 \\
& UPD [persona] & 10.969 & 1.712 & 0.101 & 0.880 & 0.061 & 0.879 & 0.696 & 0.873 \\
& UPD [context] & 10.992 & \cellcolor{blue!20}1.734 & 0.101 & 0.880 & 0.061 & 0.879 & 0.696 & 0.873 \\
& UPD [persona+context] & \cellcolor{blue!20}10.960 & 1.693 & 0.101 & 0.879 & 0.060 & 0.879 & 0.695 & 0.873 \\
& UPD [random] & 10.978 & 1.690 & 0.101 & 0.879 & 0.060 & 0.878 & 0.694 & 0.872\\
& UPR [persona] & 10.987 & 1.703 & \cellcolor{blue!20}0.102 & 0.879 & 0.060 & 0.878 & 0.695 & 0.873\\
& UPR [context] & 11.006 & 1.593 & 0.100 & 0.879 & 0.059 & 0.879 & 0.695 & 0.873\\
& UPR [persona+context] & 11.000 & 1.660 & 0.100 & 0.879 & 0.060 & 0.879 & 0.695 & 0.873 \\
& UPR [random] & 11.004 & 1.575 & 0.099 & 0.879 & 0.059 & 0.879 & 0.695 & 0.873\\
& UMD [persona] & 10.977 & 1.660 & 0.101 & 0.879 & 0.060 & 0.879 & 0.695 & 0.873\\
& UMD [context] & 11.025 & 1.659 & 0.100 & 0.879 & 0.060 & 0.879 & 0.695 & 0.873\\
& UMD [persona+context] & 11.004 & 1.714 & 0.101 & 0.879 & 0.060 & 0.879 & 0.696 & 0.873 \\
& UMD [random] & \cellcolor{blue!20}10.957 & 1.593 & 0.099 & 0.879 & 0.059 & 0.879 & 0.694 & 0.872\\
& UMR [persona] & 11.063 & 1.551 & 0.098 & 0.879 & 0.057 & 0.877 & 0.693 & 0.872 \\
& UMR [context] & 11.092 & 1.560 & 0.099 & 0.879 & 0.058 & 0.878 & 0.694 & 0.873\\
& UMR [persona+context] & 11.102 & 1.468 & 0.099 & 0.879 & 0.057 & 0.878 & 0.694 & 0.873 \\
& UMR [random] & 11.044 & 1.540 & 0.099 & 0.879 & 0.058 & 0.879 & 0.695 & 0.873\\

\bottomrule
\end{tabular}}
\caption{\label{table:main_metric} Results of the evaluation are based on automatic metrics. For each metric, colored numbers indicate the best-performing model.}
\end{table*}

\subsection{Evaluation Metrics}
The assessment of the models is performed in an automated manner utilizing well-established metrics such as perplexity \cite{vinyals_etal_2015}, BLEU \cite{papineni_etal_2002}, and Rouge-L \cite{li_etat_2016}. In addition, we also incorporate two additional methods (similarity and correlation with human judgement) for automatic evaluation. These are the Embedding Average (Average), Embedding Extrema (Extrema), and Embedding Greedy (Greedy) metrics \cite{serban_etal_2016}, which provide a deeper understanding of the correspondence between the model's responses and the reference responses. Furthermore, we compute the BertScore \cite{zhang_etal_2019} and MoverScore \cite{zhao_etall_2019} to assess the level of agreement between the generated text and human evaluations, and to determine the overall quality of the generated text.

\section{Results}
Table \ref{table:main_metric} presents the results of incorporating a combination of objectives and language modeling into various components of open-domain dialog systems. The evaluation was conducted on two benchmark datasets and two different scales of GPT-2. The results demonstrate that the improvement of the model depends on the type of auxiliary objective used in conjunction with language modeling.
As demonstrated in the table, utilizing utterance permutation in binary form has a notable impact on reducing the perplexity of the model, with a reduction of 2\% being observed. Furthermore, compared to other auxiliary tasks, the use of utterance masking in the binary form leads to improvements in metrics such as BLEU, ROUGE-L, MoverScore, and Extrema.
The results also suggest that using auxiliary tasks in larger models improves performance. The comparison between the Persona-Chat dataset highlights the significance of using auxiliary tasks simultaneously in both the Persona and Context components, which results in even better performance. Tables ~
\ref{table:dd_outputs} and \ref{table:pc_outputs}, located in Appendices \ref{sec:dd_outputs} and \ref{sec:pc_outputs} respectively, present generated responses for the two benchmarks, encompassing both the baseline and the optimal auxiliary model.\\

\noindent\textbf{What is the difference between binary and non-binary auxiliary tasks?} The results clearly demonstrate that the model only has access to the left context due to the specific type of attention mechanism employed in GPT (Masked Causal Attention). This limited exposure to context makes it challenging for the model to identify the distorted token correctly. Despite having access to the left context, the model's ability to recognize the scrambled token remains impaired. \\

\noindent\textbf{What is the impact of implementing these auxiliary tasks on different components of dialogue?} Determining the exact part of dialogue that will benefit the most from these tasks is challenging, but it can be agreed upon that combining both the Persona and Context components leads to improved outcomes. \\

\noindent\textbf{Why do the results vary across these two datasets?} The difference can be attributed to the distinct structures of the two benchmarks, as one provides access only to the context and the other to both persona and context. \\

\noindent\textbf{Does access to both persona and context result in higher quality answers?} This depends on the degree to which the persona aligns with the context.

\section{Conclusion}
In conclusion, our research has focused on improving the quality of generated responses using GPT-2 by proposing two auxiliary tasks. The first task, referred to as utterance permutation, aims to enhance the model's ability to comprehend the interconnections between words in a sentence and produce grammatically accurate responses. The second task, utterance masking, is designed to improve the coherence and consistency of the generated responses by challenging the model to predict masked words based on the surrounding context. Our experiments indicate that combining these two auxiliary tasks substantially improves the quality of generated responses. This includes improved grammar, coherence, and consistency, which are crucial aspects of generating high-quality NLP responses. Furthermore, these results demonstrate the potential of incorporating auxiliary tasks with language models to enhance their performance and improve the quality of generated responses in NLP applications.


\section*{Acknowledgements}
It is with deep gratitude that we thank the anonymous reviewers for their valuable feedback and guidance, which contributed significantly to the quality and clarity of our work. Finally, this research endeavor would not have been possible without the generous funding of the Wallenberg AI, Autonomous Systems and Software Program (WASP), funded by the Knut and Alice Wallenberg Foundation.

We would like to extend our appreciation to the Swedish National Infrastructure for Computing (SNIC), whose latest technologies enabled the computational aspects of our research. Furthermore, we acknowledge the Swedish Research Council's support for SNIC through grant agreement no. 2022/22-1003, which facilitated this research in part.

\bibliography{main}

\begin{thebibliography}{31}
\expandafter\ifx\csname natexlab\endcsname\relax\def\natexlab#1{#1}\fi

\bibitem[{Adiwardana et~al.(2020)Adiwardana, Luong, So, Hall, Fiedel,
  Thoppilan, Yang, Kulshreshtha, Nemade, Lu, and Le}]{freitas_etal_2020}
Daniel Adiwardana, Minh-Thang Luong, David~R. So, Jamie Hall, Noah Fiedel,
  Romal Thoppilan, Zi~Yang, Apoorv Kulshreshtha, Gaurav Nemade, Yifeng Lu, and
  Quoc~V. Le. 2020.
\newblock \href {http://arxiv.org/abs/2001.09977} {Towards a human-like
  open-domain chatbot}.

\bibitem[{Budzianowski and Vuli{\'c}(2019)}]{budzianowski_&_vulic_2019}
Pawe{\l} Budzianowski and Ivan Vuli{\'c}. 2019.
\newblock \href {https://doi.org/10.18653/v1/D19-5602} {Hello, it{'}s {GPT}-2 -
  how can {I} help you? towards the use of pretrained language models for
  task-oriented dialogue systems}.
\newblock In \emph{Proceedings of the 3rd Workshop on Neural Generation and
  Translation}, pages 15--22, Hong Kong. Association for Computational
  Linguistics.

\bibitem[{Budzianowski et~al.(2018)Budzianowski, Wen, Tseng, Casanueva, Ultes,
  Ramadan, and Ga{\v{s}}i{\'c}}]{budzianowski_etal_2018}
Pawe{\l} Budzianowski, Tsung-Hsien Wen, Bo-Hsiang Tseng, I{\~n}igo Casanueva,
  Stefan Ultes, Osman Ramadan, and Milica Ga{\v{s}}i{\'c}. 2018.
\newblock \href {https://doi.org/10.18653/v1/D18-1547} {{M}ulti{WOZ} - a
  large-scale multi-domain {W}izard-of-{O}z dataset for task-oriented dialogue
  modelling}.
\newblock In \emph{Proceedings of the 2018 Conference on Empirical Methods in
  Natural Language Processing}, pages 5016--5026, Brussels, Belgium.
  Association for Computational Linguistics.

\bibitem[{Chao and Lane(2019)}]{chao_&_lane_2019}
Guan-Lin Chao and Ian Lane. 2019.
\newblock \href {http://arxiv.org/abs/1907.03040} {{BERT-DST}: Scalable
  end-to-end dialogue state tracking with bidirectional encoder representations
  from transformer}.

\bibitem[{Devlin et~al.(2019)Devlin, Chang, Lee, and
  Toutanova}]{devlin_etal_2019}
Jacob Devlin, Ming-Wei Chang, Kenton Lee, and Kristina Toutanova. 2019.
\newblock \href {https://doi.org/10.18653/v1/N19-1423} {{BERT}: Pre-training of
  deep bidirectional transformers for language understanding}.
\newblock In \emph{Proceedings of the 2019 Conference of the North {A}merican
  Chapter of the Association for Computational Linguistics: Human Language
  Technologies, Volume 1 (Long and Short Papers)}, pages 4171--4186,
  Minneapolis, Minnesota. Association for Computational Linguistics.

\bibitem[{Dinan et~al.(2020)Dinan, Logacheva, Malykh, Miller, Shuster, Urbanek,
  Kiela, Szlam, Serban, Lowe, Prabhumoye, Black, Rudnicky, Williams, Pineau,
  Burtsev, and Weston}]{dinan_etal_2019}
Emily Dinan, Varvara Logacheva, Valentin Malykh, Alexander Miller, Kurt
  Shuster, Jack Urbanek, Douwe Kiela, Arthur Szlam, Iulian Serban, Ryan Lowe,
  Shrimai Prabhumoye, Alan~W. Black, Alexander Rudnicky, Jason Williams, Joelle
  Pineau, Mikhail Burtsev, and Jason Weston. 2020.
\newblock \href {https://link.springer.com/chapter/10.1007/978-3-030-29135-8_7}
  {The second conversational intelligence challenge ({ConvAI2})}.
\newblock In \emph{The NeurIPS '18 Competition}, pages 187--208, Cham. Springer
  International Publishing.

\bibitem[{Gao et~al.(2018)Gao, Galley, and Li}]{gao_etal_2018}
Jianfeng Gao, Michel Galley, and Lihong Li. 2018.
\newblock \href {https://doi.org/10.1145/3209978.3210183} {Neural approaches to
  conversational ai}.
\newblock In \emph{The 41st International ACM SIGIR Conference on Research \&
  Development in Information Retrieval}, SIGIR '18, page 1371–1374, New York,
  NY, USA. Association for Computing Machinery.

\bibitem[{Ghazvininejad et~al.(2018)Ghazvininejad, Brockett, Chang, Dolan, Gao,
  Yih, and Galley}]{ghaznavinejad_etal_2018}
Marjan Ghazvininejad, Chris Brockett, Ming-Wei Chang, Bill Dolan, Jianfeng Gao,
  Wen-tau Yih, and Michel Galley. 2018.
\newblock \href {https://doi.org/10.1609/aaai.v32i1.11977} {A
  knowledge-grounded neural conversation model}.
\newblock \emph{Proceedings of the AAAI Conference on Artificial Intelligence},
  32(1).

\bibitem[{Gu et~al.(2021)Gu, Yoo, and Ha}]{gu_etal_2020}
Xiaodong Gu, Kang~Min Yoo, and Jung-Woo Ha. 2021.
\newblock \href {https://doi.org/10.1609/aaai.v35i14.17527} {{DialogBERT}:
  Discourse-aware response generation via learning to recover and rank
  utterances}.
\newblock \emph{Proceedings of the AAAI Conference on Artificial Intelligence},
  35(14):12911--12919.

\bibitem[{Lewis et~al.(2020)Lewis, Liu, Goyal, Ghazvininejad, Mohamed, Levy,
  Stoyanov, and Zettlemoyer}]{lewis_etal_2020}
Mike Lewis, Yinhan Liu, Naman Goyal, Marjan Ghazvininejad, Abdelrahman Mohamed,
  Omer Levy, Veselin Stoyanov, and Luke Zettlemoyer. 2020.
\newblock \href {https://doi.org/10.18653/v1/2020.acl-main.703} {{BART}:
  Denoising sequence-to-sequence pre-training for natural language generation,
  translation, and comprehension}.
\newblock In \emph{Proceedings of the 58th Annual Meeting of the Association
  for Computational Linguistics}, pages 7871--7880, Online. Association for
  Computational Linguistics.

\bibitem[{Li et~al.(2016{\natexlab{a}})Li, Galley, Brockett, Gao, and
  Dolan}]{li_etat_2016}
Jiwei Li, Michel Galley, Chris Brockett, Jianfeng Gao, and Bill Dolan.
  2016{\natexlab{a}}.
\newblock \href {https://doi.org/10.18653/v1/N16-1014} {A diversity-promoting
  objective function for neural conversation models}.
\newblock In \emph{Proceedings of the 2016 Conference of the North {A}merican
  Chapter of the Association for Computational Linguistics: Human Language
  Technologies}, pages 110--119, San Diego, California. Association for
  Computational Linguistics.

\bibitem[{Li et~al.(2016{\natexlab{b}})Li, Galley, Brockett, Spithourakis, Gao,
  and Dolan}]{li_etal_2016b}
Jiwei Li, Michel Galley, Chris Brockett, Georgios Spithourakis, Jianfeng Gao,
  and Bill Dolan. 2016{\natexlab{b}}.
\newblock \href {https://doi.org/10.18653/v1/P16-1094} {A persona-based neural
  conversation model}.
\newblock In \emph{Proceedings of the 54th Annual Meeting of the Association
  for Computational Linguistics (Volume 1: Long Papers)}, pages 994--1003,
  Berlin, Germany. Association for Computational Linguistics.

\bibitem[{Li et~al.(2017)Li, Su, Shen, Li, Cao, and Niu}]{li_etal_2017}
Yanran Li, Hui Su, Xiaoyu Shen, Wenjie Li, Ziqiang Cao, and Shuzi Niu. 2017.
\newblock \href {https://aclanthology.org/I17-1099} {{D}aily{D}ialog: A
  manually labelled multi-turn dialogue dataset}.
\newblock In \emph{Proceedings of the Eighth International Joint Conference on
  Natural Language Processing (Volume 1: Long Papers)}, pages 986--995, Taipei,
  Taiwan. Asian Federation of Natural Language Processing.

\bibitem[{Loshchilov and Hutter(2017)}]{loshchilov_etal_2017}
Ilya Loshchilov and Frank Hutter. 2017.
\newblock \href {https://arxiv.org/abs/1711.05101} {Decoupled weight decay
  regularization}.
\newblock In \emph{International Conference on Learning Representations}.

\bibitem[{Luan et~al.(2017)Luan, Brockett, Dolan, Gao, and
  Galley}]{luan_etal_2017}
Yi~Luan, Chris Brockett, Bill Dolan, Jianfeng Gao, and Michel Galley. 2017.
\newblock \href {https://aclanthology.org/I17-1061} {Multi-task learning for
  speaker-role adaptation in neural conversation models}.
\newblock In \emph{Proceedings of the Eighth International Joint Conference on
  Natural Language Processing (Volume 1: Long Papers)}, pages 605--614, Taipei,
  Taiwan. Asian Federation of Natural Language Processing.

\bibitem[{Mehri et~al.(2019)Mehri, Razumovskaia, Zhao, and
  Eskenazi}]{mehri_etal_2019}
Shikib Mehri, Evgeniia Razumovskaia, Tiancheng Zhao, and Maxine Eskenazi. 2019.
\newblock \href {https://doi.org/10.18653/v1/P19-1373} {Pretraining methods for
  dialog context representation learning}.
\newblock In \emph{Proceedings of the 57th Annual Meeting of the Association
  for Computational Linguistics}, pages 3836--3845, Florence, Italy.
  Association for Computational Linguistics.

\bibitem[{Moon et~al.(2019)Moon, Shah, Kumar, and Subba}]{moon_etal_2019}
Seungwhan Moon, Pararth Shah, Anuj Kumar, and Rajen Subba. 2019.
\newblock \href {https://doi.org/10.18653/v1/P19-1081} {{O}pen{D}ial{KG}:
  Explainable conversational reasoning with attention-based walks over
  knowledge graphs}.
\newblock In \emph{Proceedings of the 57th Annual Meeting of the Association
  for Computational Linguistics}, pages 845--854, Florence, Italy. Association
  for Computational Linguistics.

\bibitem[{Papineni et~al.(2002)Papineni, Roukos, Ward, and
  Zhu}]{papineni_etal_2002}
Kishore Papineni, Salim Roukos, Todd Ward, and Wei-Jing Zhu. 2002.
\newblock \href {https://doi.org/10.3115/1073083.1073135} {{B}leu: a method for
  automatic evaluation of machine translation}.
\newblock In \emph{Proceedings of the 40th Annual Meeting of the Association
  for Computational Linguistics}, pages 311--318, Philadelphia, Pennsylvania,
  USA. Association for Computational Linguistics.

\bibitem[{Radford et~al.(2019)Radford, Wu, Child, Luan, Amodei, and
  Sutskever}]{radford_etal_2019}
Alec Radford, Jeff Wu, Rewon Child, David Luan, Dario Amodei, and Ilya
  Sutskever. 2019.
\newblock \href
  {https://d4mucfpksywv.cloudfront.net/better-language-models/language_models_are_unsupervised_multitask_learners.pdf}
  {Language models are unsupervised multitask learners}.

\bibitem[{Raffel et~al.(2020)Raffel, Shazeer, Roberts, Lee, Narang, Matena,
  Zhou, Li, and Liu}]{raffel_etal_2019}
Colin Raffel, Noam Shazeer, Adam Roberts, Katherine Lee, Sharan Narang, Michael
  Matena, Yanqi Zhou, Wei Li, and Peter~J. Liu. 2020.
\newblock \href {http://arxiv.org/abs/1910.10683} {Exploring the limits of
  transfer learning with a unified text-to-text transformer}.

\bibitem[{Sankar et~al.(2019)Sankar, Subramanian, Pal, Chandar, and
  Bengio}]{sankar_etal_2019}
Chinnadhurai Sankar, Sandeep Subramanian, Chris Pal, Sarath Chandar, and Yoshua
  Bengio. 2019.
\newblock \href {https://doi.org/10.18653/v1/P19-1004} {Do neural dialog
  systems use the conversation history effectively? an empirical study}.
\newblock In \emph{Proceedings of the 57th Annual Meeting of the Association
  for Computational Linguistics}, pages 32--37, Florence, Italy. Association
  for Computational Linguistics.

\bibitem[{Serban et~al.(2017)Serban, Sordoni, Lowe, Charlin, Pineau, Courville,
  and Bengio}]{serban_etal_2016}
Iulian Serban, Alessandro Sordoni, Ryan Lowe, Laurent Charlin, Joelle Pineau,
  Aaron Courville, and Yoshua Bengio. 2017.
\newblock \href {https://doi.org/10.1609/aaai.v31i1.10983} {A hierarchical
  latent variable encoder-decoder model for generating dialogues}.
\newblock \emph{Proceedings of the AAAI Conference on Artificial Intelligence},
  31(1).

\bibitem[{Shuster et~al.(2022)Shuster, Xu, Komeili, Ju, Smith, Roller, Ung,
  Chen, Arora, Lane, Behrooz, Ngan, Poff, Goyal, Szlam, Boureau, Kambadur, and
  Weston}]{shuster_etal_2022}
Kurt Shuster, Jing Xu, Mojtaba Komeili, Da~Ju, Eric~Michael Smith, Stephen
  Roller, Megan Ung, Moya Chen, Kushal Arora, Joshua Lane, Morteza Behrooz,
  William Ngan, Spencer Poff, Naman Goyal, Arthur Szlam, Y-Lan Boureau, Melanie
  Kambadur, and Jason Weston. 2022.
\newblock \href {http://arxiv.org/abs/2208.03188} {Blenderbot 3: a deployed
  conversational agent that continually learns to responsibly engage}.

\bibitem[{Thoppilan et~al.(2022)Thoppilan, Freitas, Hall, Shazeer,
  Kulshreshtha, Cheng, Jin, Bos, Baker, Du, Li, Lee, Zheng, Ghafouri, Menegali,
  Huang, Krikun, Lepikhin, Qin, Chen, Xu, Chen, Roberts, Bosma, Zhao, Zhou,
  Chang, Krivokon, Rusch, Pickett, Srinivasan, Man, Meier-Hellstern, Morris,
  Doshi, Santos, Duke, Soraker, Zevenbergen, Prabhakaran, Diaz, Hutchinson,
  Olson, Molina, Hoffman-John, Lee, Aroyo, Rajakumar, Butryna, Lamm, Kuzmina,
  Fenton, Cohen, Bernstein, Kurzweil, Aguera-Arcas, Cui, Croak, Chi, and
  Le}]{thoppilan_etal_2022}
Romal Thoppilan, Daniel~De Freitas, Jamie Hall, Noam Shazeer, Apoorv
  Kulshreshtha, Heng-Tze Cheng, Alicia Jin, Taylor Bos, Leslie Baker, Yu~Du,
  YaGuang Li, Hongrae Lee, Huaixiu~Steven Zheng, Amin Ghafouri, Marcelo
  Menegali, Yanping Huang, Maxim Krikun, Dmitry Lepikhin, James Qin, Dehao
  Chen, Yuanzhong Xu, Zhifeng Chen, Adam Roberts, Maarten Bosma, Vincent Zhao,
  Yanqi Zhou, Chung-Ching Chang, Igor Krivokon, Will Rusch, Marc Pickett,
  Pranesh Srinivasan, Laichee Man, Kathleen Meier-Hellstern, Meredith~Ringel
  Morris, Tulsee Doshi, Renelito~Delos Santos, Toju Duke, Johnny Soraker, Ben
  Zevenbergen, Vinodkumar Prabhakaran, Mark Diaz, Ben Hutchinson, Kristen
  Olson, Alejandra Molina, Erin Hoffman-John, Josh Lee, Lora Aroyo, Ravi
  Rajakumar, Alena Butryna, Matthew Lamm, Viktoriya Kuzmina, Joe Fenton, Aaron
  Cohen, Rachel Bernstein, Ray Kurzweil, Blaise Aguera-Arcas, Claire Cui,
  Marian Croak, Ed~Chi, and Quoc Le. 2022.
\newblock \href {http://arxiv.org/abs/2201.08239} {Lamda: Language models for
  dialog applications}.

\bibitem[{Vaswani et~al.(2017)Vaswani, Shazeer, Parmar, Uszkoreit, Jones,
  Gomez, Kaiser, and Polosukhin}]{vaswani_etal_2017}
Ashish Vaswani, Noam Shazeer, Niki Parmar, Jakob Uszkoreit, Llion Jones,
  Aidan~N Gomez, \L~ukasz Kaiser, and Illia Polosukhin. 2017.
\newblock \href
  {https://proceedings.neurips.cc/paper_files/paper/2017/file/3f5ee243547dee91fbd053c1c4a845aa-Paper.pdf}
  {Attention is all you need}.
\newblock In \emph{Advances in Neural Information Processing Systems},
  volume~30. Curran Associates, Inc.

\bibitem[{Vinyals and Le(2015)}]{vinyals_etal_2015}
Oriol Vinyals and Quoc~V. Le. 2015.
\newblock \href {https://arxiv.org/pdf/1506.05869.pdf} {A neural conversational
  model}.
\newblock In \emph{Deep Learning Workshop at ICML 2015}.

\bibitem[{Wolf et~al.(2019)Wolf, Sanh, Chaumond, and Delangue}]{wolf_etal_219}
Thomas Wolf, Victor Sanh, Julien Chaumond, and Clement Delangue. 2019.
\newblock \href {https://arxiv.org/pdf/1901.08149.pdf} {Transfertransfo: A
  transfer learning approach for neural network based conversational agents}.
\newblock \emph{ArXiv}, abs/1901.08149.

\bibitem[{Zhang et~al.(2018)Zhang, Dinan, Urbanek, Szlam, Kiela, and
  Weston}]{zhang_etal_2018}
Saizheng Zhang, Emily Dinan, Jack Urbanek, Arthur Szlam, Douwe Kiela, and Jason
  Weston. 2018.
\newblock \href {https://doi.org/10.18653/v1/P18-1205} {Personalizing dialogue
  agents: {I} have a dog, do you have pets too?}
\newblock In \emph{Proceedings of the 56th Annual Meeting of the Association
  for Computational Linguistics (Volume 1: Long Papers)}, pages 2204--2213,
  Melbourne, Australia. Association for Computational Linguistics.

\bibitem[{Zhang et~al.(2020)Zhang, Sun, Galley, Chen, Brockett, Gao, Gao, Liu,
  and Dolan}]{zhang_etal_2019}
Yizhe Zhang, Siqi Sun, Michel Galley, Yen-Chun Chen, Chris Brockett, Xiang Gao,
  Jianfeng Gao, Jingjing Liu, and Bill Dolan. 2020.
\newblock \href {https://doi.org/10.18653/v1/2020.acl-demos.30} {{DIALOGPT} :
  Large-scale generative pre-training for conversational response generation}.
\newblock In \emph{Proceedings of the 58th Annual Meeting of the Association
  for Computational Linguistics: System Demonstrations}, pages 270--278,
  Online. Association for Computational Linguistics.

\bibitem[{Zhao et~al.(2019)Zhao, Peyrard, Liu, Gao, Meyer, and
  Eger}]{zhao_etall_2019}
Wei Zhao, Maxime Peyrard, Fei Liu, Yang Gao, Christian~M. Meyer, and Steffen
  Eger. 2019.
\newblock \href {https://doi.org/10.18653/v1/D19-1053} {{M}over{S}core: Text
  generation evaluating with contextualized embeddings and earth mover
  distance}.
\newblock In \emph{Proceedings of the 2019 Conference on Empirical Methods in
  Natural Language Processing and the 9th International Joint Conference on
  Natural Language Processing (EMNLP-IJCNLP)}, pages 563--578, Hong Kong,
  China. Association for Computational Linguistics.

\bibitem[{Zhao et~al.(2020)Zhao, Xu, and Wu}]{zhao_etal_2020}
Yufan Zhao, Can Xu, and Wei Wu. 2020.
\newblock \href {https://doi.org/10.18653/v1/2020.emnlp-main.279} {Learning a
  simple and effective model for multi-turn response generation with auxiliary
  tasks}.
\newblock In \emph{Proceedings of the 2020 Conference on Empirical Methods in
  Natural Language Processing (EMNLP)}, pages 3472--3483, Online. Association
  for Computational Linguistics.

\end{thebibliography}

\appendix
\onecolumn
\section{A comparison of Vanilla GPT-2 and GPT-2 with Auxiliary Tasks in Conversational Contexts for the DailyDialogue Dataset}
\label{sec:dd_outputs}
\begingroup
\setlength{\tabcolsep}{15pt}
\renewcommand{\arraystretch}{0}
\begin{table}[ht!]
\normalsize
\centering
\begin{adjustbox}{totalheight=\textheight-2\baselineskip-4cm}
      \resizebox{\textwidth}{!}{
\begin{tabular}{>{\centering\arraybackslash}m{0.2cm}>{\arraybackslash}m{7.0cm}>{\arraybackslash}m{3.5cm}>{\arraybackslash}m{4.0cm}}
\toprule
\midrule
\rotatebox{90}{ Model } & Conversation & Ground response & Generated response \\ \hline

\rotatebox{90}{ VanillaGPT-2 [sm] }
& 
\vspace{0.5cm}
- By the way miss, where is the toilet? \newline
- Toilets are in the rear, I am afraid all the toilets are fully occupied at the moment. \newline
- What? Oh, what we live! Thank you very much for your help, miss. \newline
& 
 You are welcome.
& 
- Thanks. \newline
- I'll see you in Beijing in a few days. \newline
- Goodbye, bye. \newline
\\ \hline

\rotatebox{90}{ UPD [sm][context] }
& 
\vspace{0.5cm}
- By the way miss, where is the toilet? \newline
- Toilets are in the rear, I am afraid all the toilets are fully occupied at the moment. \newline
- What? Oh, what we live! Thank you very much for your help, miss. \newline
& 
 You are welcome.
& 
- Don't mention it. \newline
- You are welcome. \newline
- You are welcome. \newline
\\ \hline

\rotatebox{90}{ VanillaGPT-2 [md] }
& 
\vspace{0.5cm}
- Yeah, look at all those new buildings going up! Mr. Zhang, the traffic is pretty smooth. But we were told the roads from the airport to downtown were quite crowded and traffic jams could be as long as half an hour.\newline
- Yes, they were. But it has already past. The traffic from the airport to downtown has been relieved after the completion of Yan'an Aerial Road.\newline
- We've never expected that Shanghai is such a big metropolis, same as New York city. David, would it be a good idea to rent a car here? I've got an international driver's license with a pretty good record in the States. \newline
& 
\vspace{0.5cm}
That's wonderful. OK. I'll remember that. But I have to warn you that you must drive very carefully in the downtown area, for there're no separate lanes for motor vehicles and bicycles in some places, and still quite a few people don't take road regulations seriously. \newline
& 
\vspace{0.5cm}
- Oh, yeah, you'd better have it. \newline
- Oh, no, that's stupid. You'd better drive a taxi! \newline
- That's a good idea. \newline
\\ \hline

\rotatebox{90}{ UPD [md][context] }
& 
\vspace{0.5cm}
- Yeah, look at all those new buildings going up! Mr. Zhang, the traffic is pretty smooth. But we were told the roads from the airport to downtown were quite crowded and traffic jams could be as long as half an hour.\newline
- Yes, they were. But it has already past. The traffic from the airport to downtown has been relieved after the completion of Yan'an Aerial Road.\newline
- We've never expected that Shanghai is such a big metropolis, same as New York city. David, would it be a good idea to rent a car here? I've got an international driver's license with a pretty good record in the States. \newline
& 
\vspace{0.5cm}
That's wonderful. OK. I'll remember that. But I have to warn you that you must drive very carefully in the downtown area, for there're no separate lanes for motor vehicles and bicycles in some places, and still quite a few people don't take road regulations seriously.\newline
& 
\vspace{0.5cm}
- Mr. Zhang, I can assure you that the traffic isn't that bad here. There's a traffic island at the edge of the city to ease the traffic. But I still prefer to drive my own car. \newline
- Mr. Zhang, I can assure you that the traffic isn't that bad here. There's a traffic island at the edge of the city to ease the traffic. But I still prefer to drive my own car. \newline
- That would be great! \newline
\\ \hline

\bottomrule
\end{tabular}
}
\end{adjustbox}
\caption{\small A comparison of Vanilla GPT-2 and Utterance Permutations (binary) models with small and medium scales, demonstrating ground truth responses and generated responses for various dialogues. Overall, the responses are more or less related to the context, but the GPT-2 with the auxiliary tasks produces responses that are more closely related to the context. }
\label{table:dd_outputs}
\end{table}
\endgroup
\section{A comparison of Vanilla GPT-2 and GPT-2 with Auxiliary Tasks in Conversational Contexts for the PersonaChat Dataset}
\label{sec:pc_outputs}
\begingroup
\setlength{\tabcolsep}{15pt}
\renewcommand{\arraystretch}{0}
\begin{table}[ht!]
\normalsize
\centering
\begin{adjustbox}{totalheight=\textheight-2\baselineskip-4cm}
      \resizebox{\textwidth}{!}{
\begin{tabular}{>{\centering\arraybackslash}m{0.1cm}>{\arraybackslash}m{5cm}>{\arraybackslash}m{7.0cm}>{\arraybackslash}m{3.0cm}>{\arraybackslash}m{5cm}}
\toprule
\midrule
\rotatebox{90}{ Model } & Persona & Conversation & Ground response & Generated responses \\ \hline
\rotatebox{90}{VanillaGPT-2 [sm]}
& 
- I love to drink wine and dance in the moonlight.\newline
- I am very strong for my age.\newline
- I'm 100 years old.\newline
- I feel like I might live forever.\newline
&
\vspace{0.5cm}
- Hi, how are you doing today?\newline
- Good! you! Celebrating with fellow centenarians.\newline
- Nice. I'm playing some card games with my family.\newline
- That sounds like fun. I like wine and dancing too!\newline
- Same. I try to get a small workout in a three mile walk for me is key.\newline
- Awesome! Walking like that you might live forever like me!!\newline
- That's true. My favorite way to unwind is reading a book.\newline
& 
 You sound very like minded, just add some wine.
& 
- I know! Where do you live? I'm still 100 years from retirement!\newline
- What is your favorite book? Do you like to read?\newline
- That's a big idea! I love books too.
\\ \hline

\rotatebox{90}{ VanillaGPT-2 [sm] }
& 
\vspace{0.5cm}
- I'm now looking for a new job. \newline
- My favorite movie is Friday. \newline
- I am a doctor. \newline
- My daughter is a child prodigy. \newline
&
- Hello, I'm sitting here with my dog. How are you? \newline
& 
 I'm well friend. Looking for new employment at the moment.
& 
- Hi my name is Dr. Robert. \newline
- Hi how are you today.\newline
- Hi. how are you doing?
\\ \hline

\rotatebox{90}{ UMB [sm][random] }
& 
- I love to drink wine and dance in the moonlight.\newline
- I am very strong for my age.\newline
- I'm 100 years old.\newline
- I feel like I might live forever.\newline
&
\vspace{0.5cm}
- Hi, how are you doing today?\newline
- Good! you! Celebrating with fellow centenarians.\newline
- Nice. I'm playing some card games with my family.\newline
- That sounds like fun. I like wine and dancing too!\newline
- Same. I try to get a small workout in a three mile walk for me is key.\newline
- Awesome! Walking like that you might live forever like me!!\newline
- That's true. My favorite way to unwind is reading a book.\newline
& 
 You sound very like minded, just add some wine.
& 
- That sounds like a lot of fun.\newline
Very healthy. I'm really strong. You should try to eat healthy!\newline
Books are fun. I have tons.\newline
\\ \hline

\rotatebox{90}{ UMB [sm][random] }
& 
\vspace{0.5cm}
- I'm now looking for a new job. \newline
- My favorite movie is Friday. \newline
- I am a doctor. \newline
- My daughter is a child prodigy. \newline
&
- Hello, I'm sitting here with my dog. How are you? \newline
& 
 I'm well friend. Looking for new employment at the moment.
& 
\vspace{0.5cm}
- I'm doing good. Just finished my shift at the hospital. How are you? \newline
- I'm doing well, and yourself?\newline
- I am well, how about yourself.\newline
\\ \hline

\rotatebox{90}{ VanillaGPT-2 [md] }
& 
\vspace{0.5cm}
- I prefer being inside.\newline
- I am a recording engineer and own my own studio.\newline
- Whats up I am a 46 year old single dad 46 a kids.\newline
- I live in California but the recording artist market was dead last year.\newline
&
- Hello, call of duty is my favorite game. Do you game?\newline
& 
  Yes I do after all I am a engineer.
& 
\vspace{0.5cm}
- Not as much as my daughter she's my baby.\newline
- No I am just a big time recording engineer.\newline
- I prefer the movie games and tv shows. I have a studio in California.\newline
\\ \hline

\rotatebox{90}{ UMB [md][random] }
& 
\vspace{0.5cm}
- I prefer being inside.\newline
- I am a recording engineer and own my own studio.\newline
- Whats up I am a 46 year old single dad 46 a kids.\newline
- I live in California but the recording artist market was dead last year.\newline
&
- Hello, call of duty is my favorite game. Do you game?\newline
& 
  Yes I do after all I am a engineer.
& 
\vspace{0.5cm}
- I don't play game like that. I'm a live video coordinator.\newline
- Not really. What do you do?\newline
- I am not sure I've the computer but the tv show is my favorite.\newline
\\ \hline

\bottomrule
\end{tabular}
}
\end{adjustbox}
\caption{\small An examination of Vanilla GPT-2 and Binary Utterance Masking (random in context and persona) models in small and medium sizes, showcasing authentic responses. The responses generally have a degree of contextual and persona relevance, but when the GPT-2 is integrated with auxiliary tasks, the responses demonstrate a stronger connection to the context and persona.}
\label{table:pc_outputs}
\end{table}
\endgroup

\end{document}